\documentclass[runningheads]{llncs}
\usepackage{graphicx}
\usepackage{times,slashbox}
\usepackage{graphicx,xcolor}
\usepackage{amsmath,amssymb,bbm}
\usepackage{enumitem}
\usepackage{booktabs}  % professional-quality tables
\usepackage{algorithm}
\usepackage{algpseudocode}

\def\R{{\mathbb R}}

\def\Ncal{{\mathcal N}}

\definecolor{darkblue}{rgb}{0.1,.1,0.8}
\begin{document}
\title{Diversity in deep generative models and generative AI}
\titlerunning{Diversity in generative AI}
\author{Gabriel Turinici\inst{1}
\orcidID{0000-0003-2713-006X}}
%	\author{-} 
	
%https://orcid.org/0000-0003-2713-006X
\authorrunning{G. Turinici}
%\authorrunning{-}
\institute{CEREMADE, \\  Universit\'e Paris Dauphine - PSL, CNRS, Paris, France \\
		\email{gabriel.turinici@dauphine.fr}\\
		\url{https://turinici.com} }
%\institute{- \\	\email{x.x@x.x}\\	\url{https://x.com} }
	\maketitle              % typeset the header of the contribution
\begin{abstract}
The decoder-based machine learning generative algorithms such as Generative Adversarial Networks (GAN), Variational Auto-Encoders (VAE), Transformers show impressive results when constructing objects similar to those in a training ensemble. 
However, the generation of new objects builds mainly on the understanding of the hidden structure of the training dataset followed by a sampling from a multi-dimensional normal variable. In particular each sample is independent from the others and can repeatedly propose same type of objects. To cure this drawback we introduce a kernel-based  measure quantization method that can produce new objects from a given target measure by approximating it as a whole and even staying away from elements already drawn from that distribution. This ensures a better diversity of the produced objects. The method is tested on classic machine learning benchmarks.

\end{abstract}

\section{Introduction, motivation and literature review}
\label{sec:intro}

We investigate in this work an approach to enhance diversity in decoder-based generative AI paradigms i.e., when generating objects (e.g., images) similar to the content of a given (training) dataset. Such procedures received a large audience in the last years, especially after the introduction of several deep neural network architectures widely used today: the Generative Adversarial Networks~\cite{originalGAN14,tolstikhin2017wasserstein,arjovsky2017wgan} (hereafter named GAN), the Variational Auto-Encoders (VAE), see~\cite{kingma2013autoencoding,kingma_introduction_2019,CWAE} and the Transformer \cite{vaswani_attention_2017}.

All approaches use a small-dimensional set of parameters called {\it latent space} of dimension $L$ as a companion representation for any object of initial dimension $N$ ; for instance, for RGB color pictures $N$ will be $3$ times the number of pixels. We will take VAE as an example.  
At the high level of description, an object e.g., an image, is a vector in $\R^N$; the training dataset becomes a set of points in $\R^N$ and it is hypothesized that it corresponds to some distribution $\mu(dx)$\footnote{The notation $\mu(dx)$ means that $\mu$ is a distribution of objects in $\R^N$ with 
generic variable $x$.} on $\R^N$ of which the dataset is an empirical sampling. The object distribution $\mu$ is mapped by the encoder part of the VAE  into an empirical distribution $\mu_L$ on the latent space $\R^L$ ; here $L$ is much smaller than $N$ and represents the essential degrees of freedom. The existence of such a $L$ is a crucial hypothesis of most generative AI procedures.
 The optimization routines of VAE ensure that this mapping of the training  dataset as a probability distribution on $\R^L$ 
will be as close as possible to some target (ideal) latent distribution, chosen usually to be the multi-variate normal distribution $\Ncal(0_L,\textrm{Id}_L)$~\footnote{Some other choices exist, a popular one being a mixture of normal variables \cite{dilokthanakul2017deep}.}; here $0_L$ is the zero vector in $\R^L$ and $\textrm{Id}_L$ is the identity matrix in $\R^{L \times L}$. 
The GAN and Transformer architectures operate a bit differently but in all cases, generating $J$ new objects resumes to drawing $J$ independent new samples $X_1,..., X_J$ from $\Ncal(0_L,\textrm{Id}_L)$. These samples are then ``decoded", i.e., passed through a neural network implementing a mapping 
$D : \R^L \to \R^N$ with $D(X_j)$ being the object e.g., image, corresponding to the latent representation $X_j \in \R^L$ for any $j\le J$.

Since the $J$ random variables $X_1,..., X_J$ are independent, some $X_j$ end up being very similar and the decoded objects $D(X_j)$ may lack diversity.

The goal of this work is to propose a method to enforce this diversity. We do this by relating the samples $X_j$ through the requirement that $ \frac{1}{J}\sum_{j=1}^J \delta_{X_j}$ be as close as possible to the empirical distribution $\mu_L$ on the latent space or the ideal latent  distribution $\Ncal(0_L,\textrm{Id}_L)$ ;  here $\delta_x$ is a general notation for the Dirac measures centered at the value $x$.
Such a goal will ensure that the objects $X_j$ cover well the (empirical) latent distribution $\mu_L$ and the generated objects $D(X_j)$ have adequate diversity.

\subsection{Short literature review}

Representing a target measure $\mu$ by the means of a sum of Dirac measures is similar in principle to the ``vector quantization" approaches~\cite{r_gray_vector_1984,book_quantization_measures} that divide the support of $\mu$ in several regions, called Voronoi cells, and replace the values in any Voronoi cell by the value at the center of the cell. It results a ``quantized" set of values hence the name of the method.

Our approach is similar to this method, but from a technical point of view we do not work in the 
 Wasserstein norm (which is used by the vector quantization algorithms)
 but instead our proposal is based on a kernel which gives rise to interesting analytic properties (for instance the distance to a normal can be calculated explicitly, see~\cite{turinici_radonsobolev_2021} for details).

From the computational point of view, our contribution is similar to the ``energy statistic" 
(see \cite{sriperumbudur_universality_2011,SZEKELY13}) but with the modification that we use a kernel which is not exactly $|x|$ but a smooth approximation.

On the other hand, diversity has been evoked 
in a recent work \cite{allahyani_divgan_2023} in the context of GANs when avoiding the ``mode collapse". The authors proposes a new GAN framework called diversified GAN (DivGAN) that aims at encouraging the GANs to produce diverse data. The DivGAN module computes a metric called ``contrastive loss" that indicates with the level of diversity in the sample. Their approach has objectives aligned with our but instead of the contrastive loss we use state of the art kernel-based statistical distance as in \cite{turinici_radonsobolev_2021}. Note also the approach of \cite{Liu_2020_CVPR} that uses uses conditional GANs to avoid mode collapse. 

Finally, for a more general discussion on the diversity and fidelity metrics see~\cite{pmlr-v119-naeem20a} that propose new metrics to upgrade the standard ones like the Inception Score (IS) and the Frechet Inception Distance (FID).

\section{Representation of the target distribution}
\label{sec:print}

We describe in this section the main part of our procedure. The procedure is based on the minimization, with respect to $X_j$, of the distance 
between the Dirac measure sum 
$ \frac{1}{J}\sum_{j=1}^J \delta_{X_j}$ and the target distribution  $\mu_L$ or $\Ncal(0_L,\textrm{Id}_L)$ (recall that the goal of the VAE is to render $\mu_L$ very close to $\Ncal(0_L,\textrm{Id}_L)$). We use the Adam~\cite{kingma2014adam} stochastic optimization algorithm to minimize the distance but one can also choose Nesterov, momentum, SGD etc. 

To compute the distance between two sets of Dirac measures we employ the following metric:
\begin{eqnarray}
& \ & 
d\left( \frac{1}{J}\sum_{j=1}^J \delta_{X_j}, \frac{1}{B}\sum_{b=1}^B \delta_{z_b} \right)^2
= \frac{\sum_{j,b=1}^{J,B} h(X_j-z_b) }{J B}
\nonumber \\ & \ & 
-  \frac{\sum_{j,j'=1}^{J} h(X_j-X_{j'})}{2 J^2} %\nonumber \\ & \ & 
-  \frac{\sum_{b,b'=1}^{B} h(z_b-z_{b'})}{2 B^2},
\label{eq:distanceformula}
\end{eqnarray}
where the kernel $h$ is defined for any $a\ge 0$ by 
$h(x)=\sqrt{ \|x\|^2 +a^2}- a$.
%\begin{remark}
The distance $d(\cdot,\cdot)$ is a kernel-based statistical distance; we refer to \cite{turinici_radonsobolev_2021,turinici_huber_2022} for considerations its usefulness and properties. Note that in particular it is not obvious that $d\left( \frac{1}{J}\sum_{j=1}^J \delta_{X_j}, \frac{1}{B}\sum_{b=1}^B \delta_{z_b} \right)\ge 0$ for any choice of vectors $X$ and $z$, but this can be proven with tools from the theory of Reproducing Kernel Hilbert Spaces (RKHS); in particular in \cite{turinici_huber_2022} it is proven that such a distance is of Gaussian mixture type. This statistical distance can be extended also to general measures (not only sums of Diracs) by the definition~:
\begin{equation}
d(\mu_1,\mu_2)^2= -\frac{1}{2}\int_{\R^L} \int_{\R^L} h(x-y) (\mu_1-\mu_2)(dx) (\mu_1-\mu_2)(dy).
\end{equation}

%\end{remark}

With these provisions we can introduce our two algorithms below  \ref{alg:tcalgo_sample_normal} and	\ref{alg:tcalgo_sample_empirical}.
Both use repeated sampling from the target distribution in order to stochastically minimize the distance from the candidate
$ \frac{1}{J}\sum_{j=1}^J \delta_{X_j}$ 
to the target distribution.
The difference between the two algorithms is the following:  \ref{alg:tcalgo_sample_normal} samples  from the ideal distribution $\Ncal(0_L,\textrm{Id}_L)$
while \ref{alg:tcalgo_sample_empirical} samples from the empirical distribution $\mu_L$. 

The parameters of the Adam algorithm were set to the defaults. Note that here the unknowns are coordinates of the vectors $X_j$ and the goal of the algorithm is to find the optimal $X_j$, $j=1,..,J$. To do so, the stochastic optimization algorithm needs to compute the gradient, with respect to $X$ of the loss function. Such a computation is done with the usual tools of back-propagation even if here the learned parameters do not correspond to neural network layers.

\begin{algorithm}
	\caption{Diversity sampling algorithm : ideal target case $\Ncal(0_L,\textrm{Id}_L)$}
	\label{alg:tcalgo_sample_normal}
{\bf Inputs : }  batch size $B$, parameter $a=10^{-6}$.

{\bf Outputs : } quantized points $X_j$, $j=1...,J$.

	\begin{algorithmic}[1]
		\Procedure{}{}%{batch size $K$, latent dimension $N$}
%		\State $\bullet$ {\bf inputs : } batch size $B$, parameter $a=10^{-6}$
		\State initialize points $X=(X_j)_{j=1}^J$ sampled i.i.d from $\Ncal(0_L,\textrm{Id}_L)$;
		\While{(max iteration not reached)}
		\State  sample i.i.d $z_1,...,z_B \sim \Ncal(0_L,\textrm{Id}_L)$;
		\State  compute the global loss	
		$L(X) := d\left( \frac{1}{J}\sum_{j=1}^J \delta_{X_j}, 
		\frac{1}{B}\sum_{b=1}^B \delta_{z_b} \right)^2$ as in eq. \eqref{eq:distanceformula}~;		
		\State  update $X$ by performing one step of the Adam algorithm to minimize $L(X)$.	
		\EndWhile\label{euclidendwhile}
		\EndProcedure
	\end{algorithmic}
\end{algorithm}

%\begin{remark}
Note that, even if we describe the measure representation for the particular situation that have as target a multi-dimensional normal distribution, the procedure above can be generalized to any other targets. On the other hand, for the normal distribution, the distance from a sum of Dirac measures to the normal distribution can be computed analytically as in \cite{turinici_radonsobolev_2021}; such an analytic formula renders the  sampling from the target distribution 
useless and the whole stochastic optimization in algorithm \ref{alg:tcalgo_sample_normal} can be replaced by a deterministic optimization; we tested the procedure and the results were coherent with what is reported here.
%\end{remark}

\begin{algorithm}
	\caption{Diversity sampling algorithm : empirical target $\mu_L$}
	\label{alg:tcalgo_sample_empirical}
	{\bf Inputs : }  batch size $B$, parameter $a=10^{-6}$, measure $\mu_L$ stored previously or computed on the fly.
	
	{\bf Outputs : } quantized points $X_j$, $j=1...,J$.
	
	\begin{algorithmic}[1]
		\Procedure{}{}%{batch size $K$, latent dimension $N$}
		%		\State $\bullet$ {\bf inputs : } batch size $B$, parameter $a=10^{-6}$
		\State initialize points $X=(X_j)_{j=1}^J$ sampled i.i.d from $\mu_L$;
		\While{(max iteration not reached)}
		\State  sample i.i.d. $z_1,...,z_B \sim \mu_L$;
		\State  compute the global loss		
		$L(X) := d\left( \frac{1}{J}\sum_{j=1}^J \delta_{X_j}, 
		\frac{1}{B}\sum_{b=1}^B \delta_{z_b} \right)^2$ as in eq. \eqref{eq:distanceformula}~;		
		\State update $X$ by performing one step of the Adam algorithm to minimize $L(X)$.			\EndWhile\label{euclidendwhile_empirical}
		\EndProcedure
	\end{algorithmic}
\end{algorithm}

The algorithm \ref{alg:tcalgo_sample_empirical} is tailored specifically for VAE. It samples from the empirical distribution $\mu_L$ in several possible distinct manners:
\begin{enumerate}
\item  store, during the last epoch of the VAE convergence, the latent points and construct $\mu_L$ as a list of Dirac masses. There is no additional computation cost but memory is required to store the data; memory consumption is usually not large because the latent space has very reduced dimension compared to the initial dataset;
\item \label{item:recompute_latent}
as previously but the computation is done {\bf after } the last epoch of the GAN / VAE / Transformer ; the computational cost increases with less than the cost of one additional epoch of the algorithm
(no decoding and no need to compute gradients);
\item on the fly: when sampling from $\mu_L$ is required, select at random objects from the initial dataset and encode them into the latent space. The cost is that of the encoding step.
\end{enumerate}

In practice in the numerical tests we selected alternative \ref{item:recompute_latent} which gives best quality at a very reasonable cost.

\section{Numerical results for the ideal sampling algorithm \ref{alg:tcalgo_sample_normal}}
\label{sec:numerics_ideal}

All the experiments below are available on the Github site \cite{turinici_huber_2022} and also as Zenodo repository \cite{turinici_supporting_2023}. In order to test both algorithm we used, even for the  algorithm \ref{alg:tcalgo_sample_normal}, a VAE setting.

\subsection{The VAE design}

\begin{figure*}
	\begin{center}
		\includegraphics[width=0.75\linewidth]{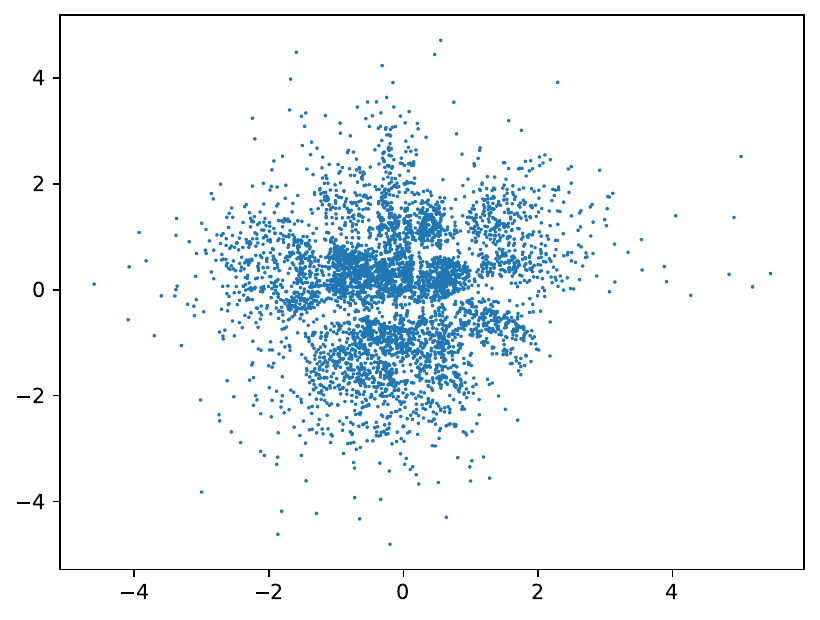}
		\caption{The latent space representation of the MNIST dataset; we used $5000$ images in the dataset. Each image is encoded and its corresponding 2D latent vector is plotted. Compare with figure \ref{fig:2Dvisualizationmnist} that displays some decoded images; the latent distribution is close to a 2D Gaussian but is not fully so.}  \label{fig:2Dlatentmnist}
	\end{center}
\end{figure*} 

\begin{figure*}
	\begin{center}
		\includegraphics[width=0.75\linewidth]{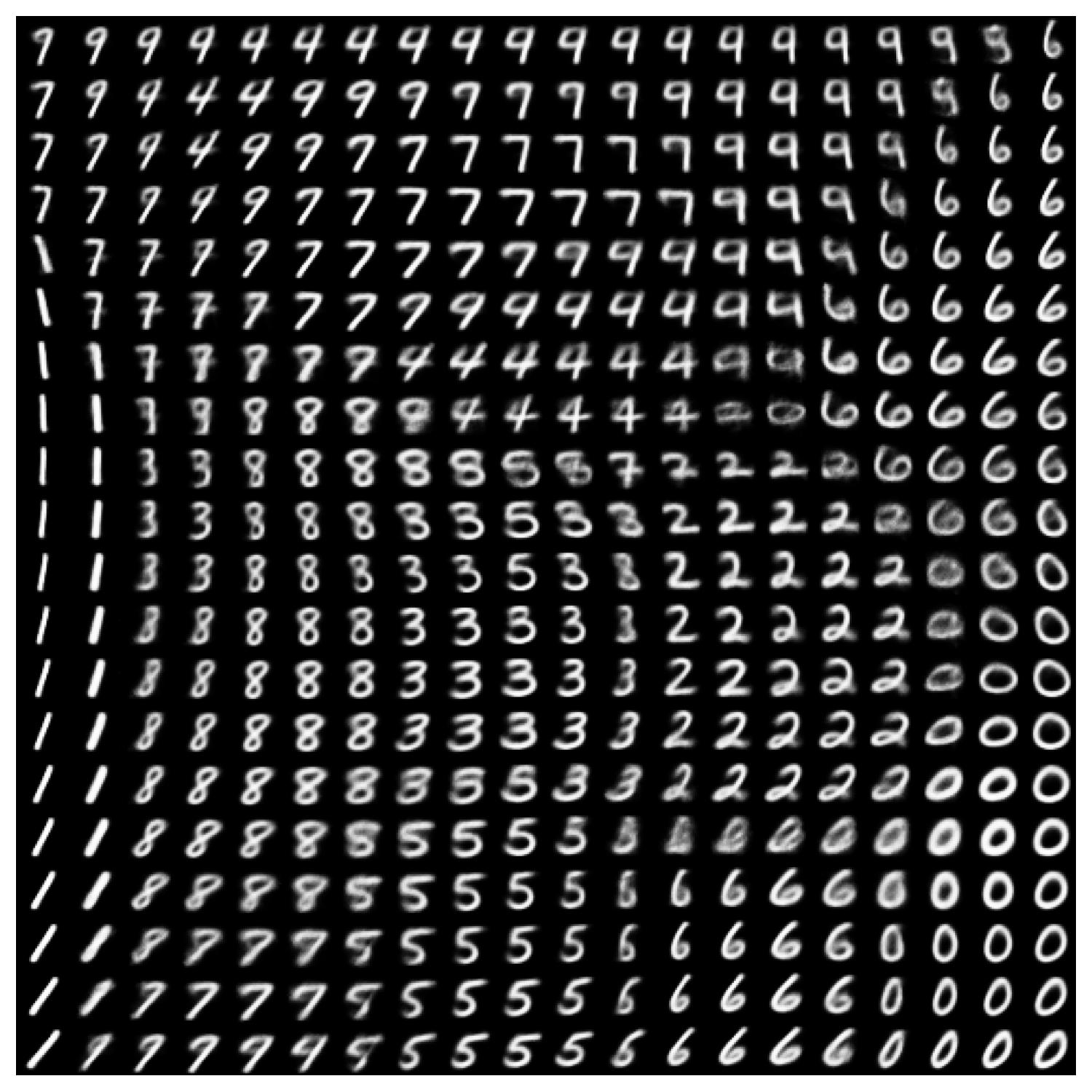}
		\caption{The latent space representation of the MNIST dataset; we use the same approach as in \cite{cvae_tensorflow_jan22} and sample the distribution with $Q=20$ equidistant (quantile-wise) points, for instance the point in the lattice at line $i_1$ and column $i_2$ corresponds to the $i_1/Q$-th quantile in the first dimension and  $i_2/Q$-th quantile in the second direction (for the normal distribution). For each such a point we draw  the image associated by the decoder $D(\cdot)$ to that point. 
		}  \label{fig:2Dvisualizationmnist}
	\end{center}
\end{figure*}

We take as example the MNIST dataset (similar results, not shown here, were obtained for the Fashion-MNIST dataset~\cite{fmnist_dataset}) and generate new images through a VAE; more precisely
we use a standard VAE which is the CVAE in the Tensorflow tutorial \cite{cvae_tensorflow_jan22}~; 
however, in order to gain in quality, we replace all convolution Conv2D layers by fully connected (FC) layers (size $28*28$)
 which results in the following encoder / decoder architecture:

\noindent{\bf Encoder}: input $28\times 28$ images; followed by $5$ 
 Relu FC layers of dimension $28*28$ and a final dense layer of dimension $2L$ (no activation).

\noindent{\bf Decoder}:
$4$ Relu FC layers of dimension $28*28$ and a final dense layer of dimension $28*28$ (no activation).

The latent space dimension is $L=2$; the encoding mapping with respect to the image dataset is presented in figures \ref{fig:2Dlatentmnist} and \ref{fig:2Dvisualizationmnist} where a good quality is observed, even if some figures, 
like the $4$ and $3$ are not well represented (all $4$ resemble very much to a $9$). Note that although the latent space distribution is close to a 2D Gaussian it is not exactly so. This will affect the quality of the generated images which is not yet optimal.

\subsection{Diversity enforcing sampling}

\begin{figure*}
	\begin{center}
		\includegraphics[width=0.75\linewidth]{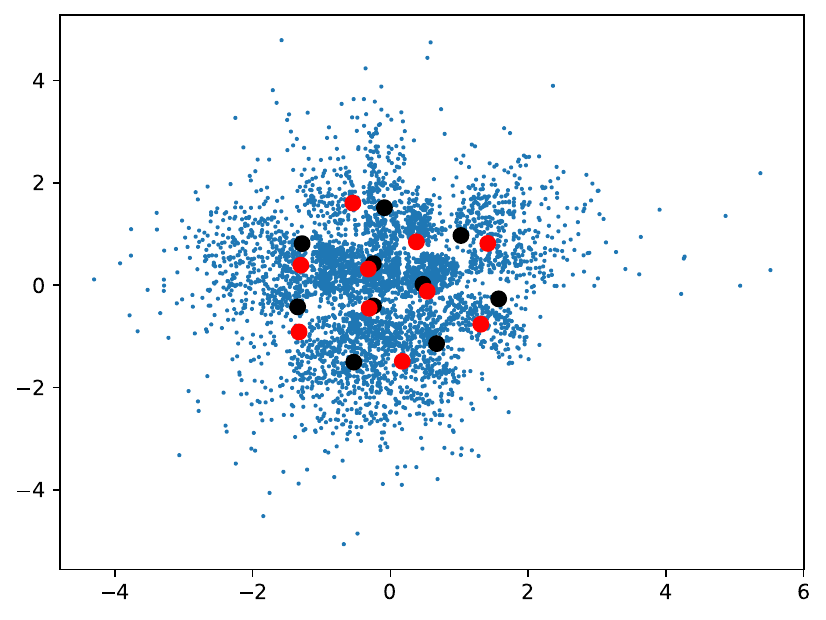}
		\caption{The latent space representation of the MNIST dataset together with the two sets of latent points corresponding to diversity sampling depicted in the second row of figure
	 \ref{fig:comparison}. Blue points are latent distribution points $\mu_L$; red and 
	 black points are the two sets of results $X=(X_j)_{j=1}^J$ of the two runs of the algorithm
	   \ref{alg:tcalgo_sample_normal} (red =first run, black= second run).
	 }  \label{fig:2Dlatentmnist_plus_points}
	\end{center}
\end{figure*}

\begin{figure*}
	\begin{center}
		\includegraphics[width=0.3\linewidth]{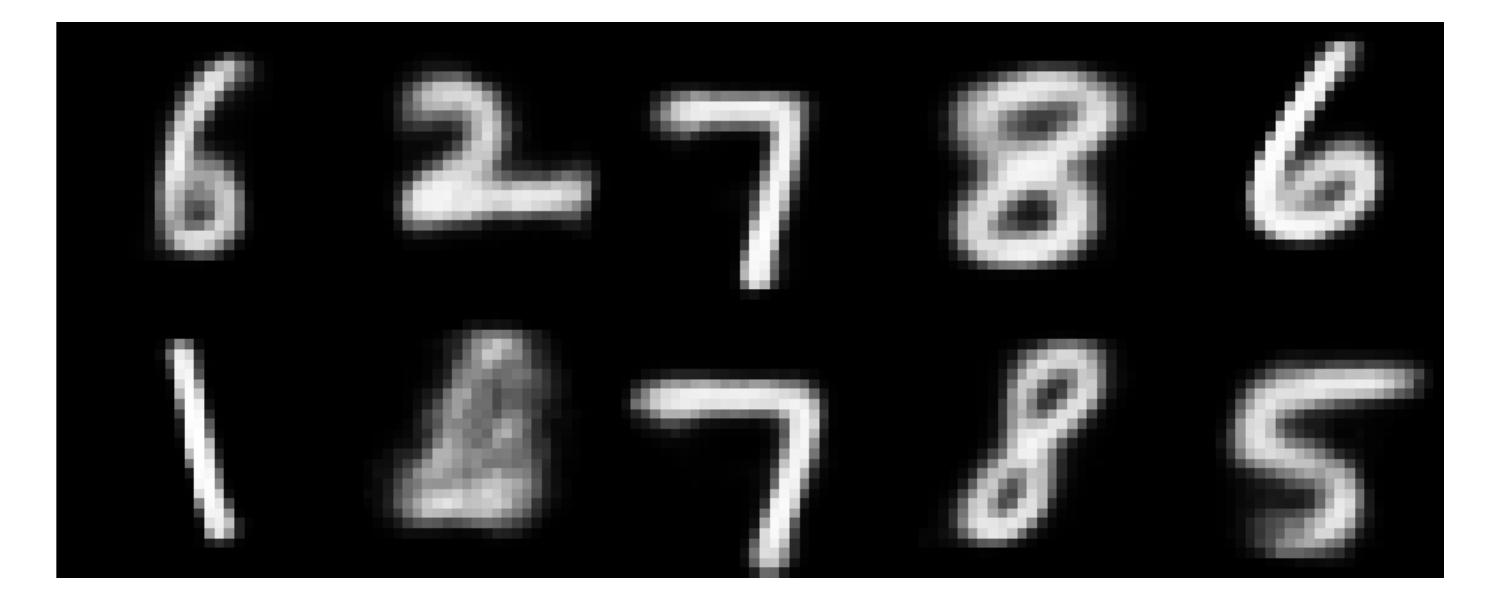}
		\includegraphics[width=0.3\linewidth]{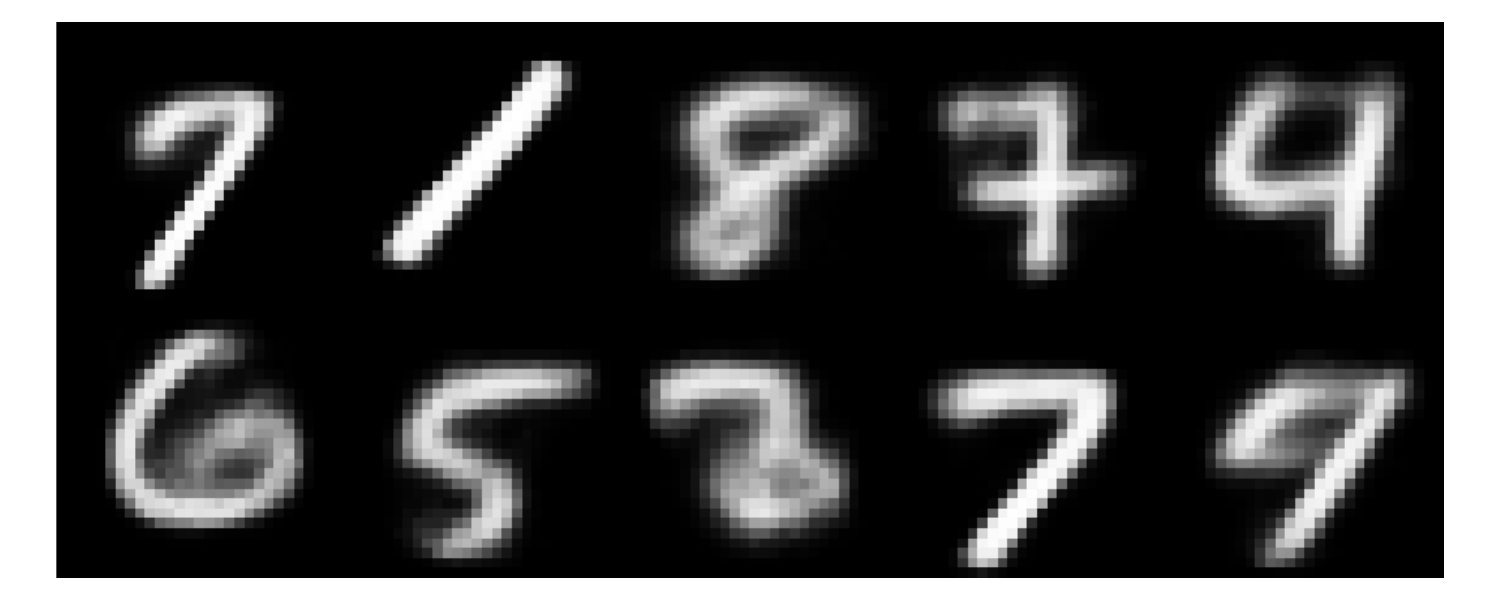}
\vspace{-0.6cm}
\begin{center}
%\hline
\rule{0.6\linewidth}{0.1pt}
\end{center}
\vspace{-0.2cm}

\includegraphics[width=0.3\linewidth]{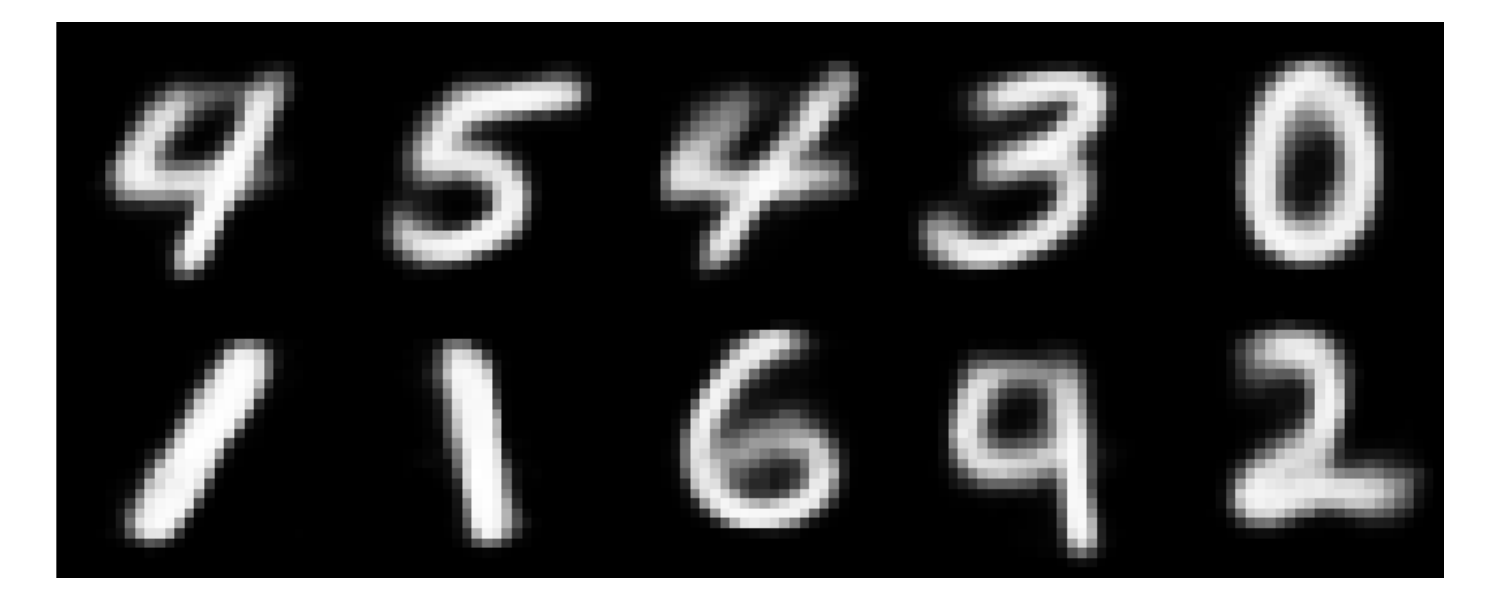}
\includegraphics[width=0.3\linewidth]{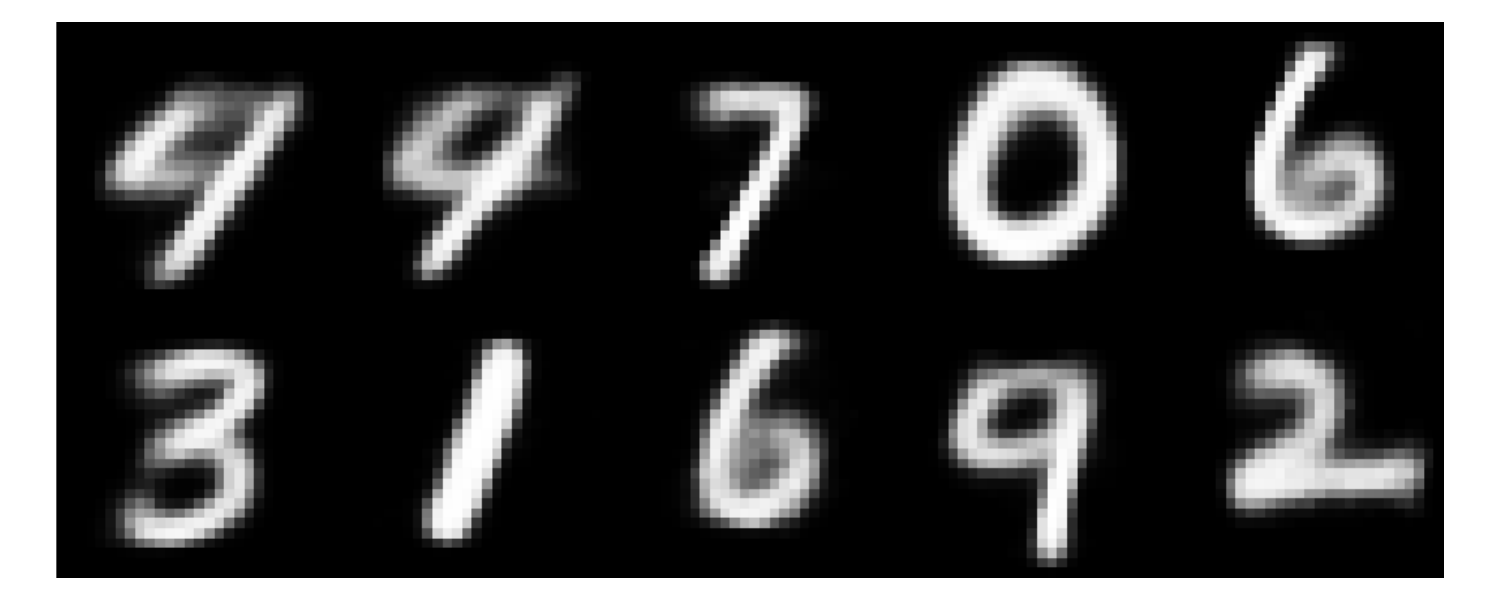}

\includegraphics[width=0.3\linewidth]{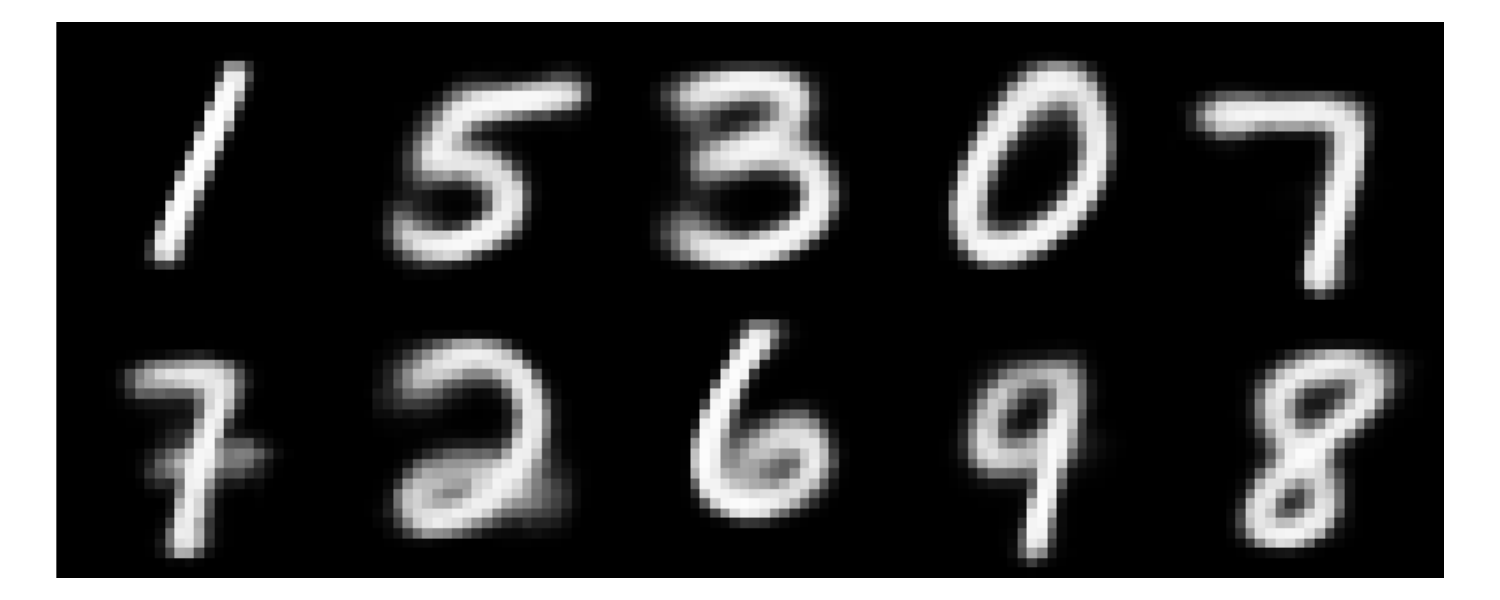}
\includegraphics[width=0.3\linewidth]{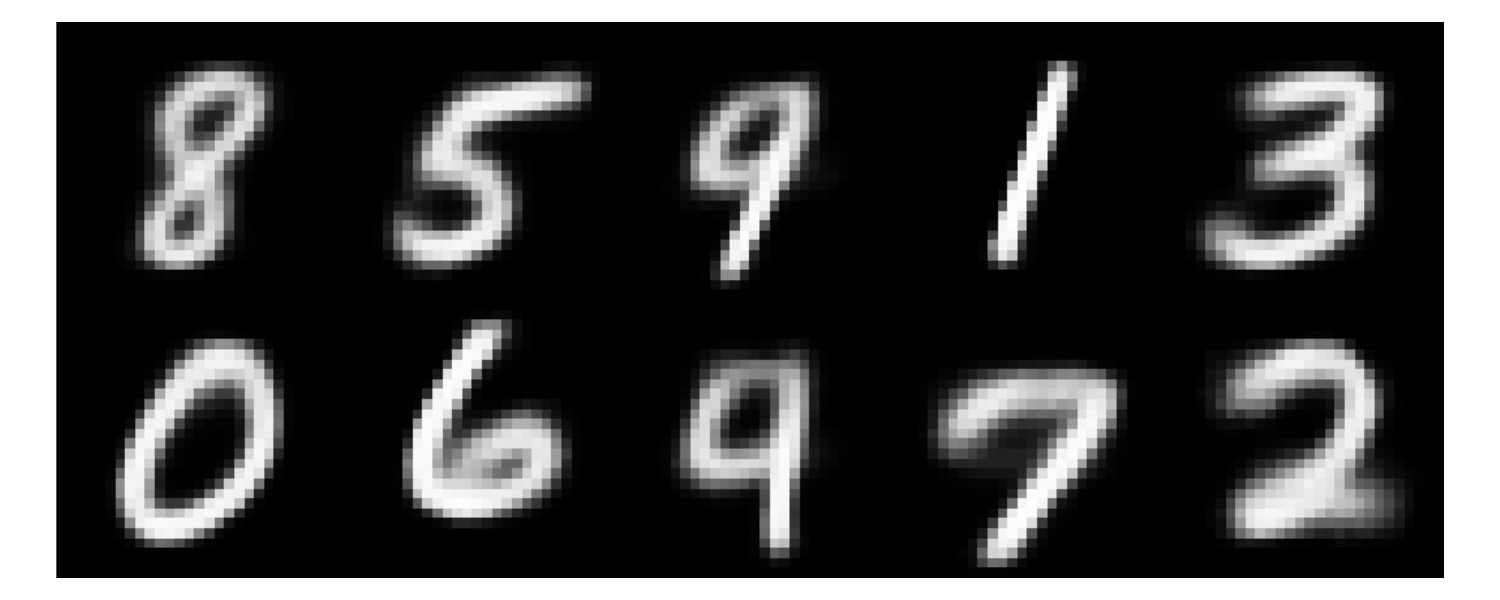}

\caption{Diversity sampling results from algorithms \ref{alg:tcalgo_sample_normal} and \ref{alg:tcalgo_sample_empirical}. 
{\bf First row pictures~:} I.i.d. sampling of $J=10$ 
points from the target latent distribution (2D normal) and their corresponding images (after decoding); we took two independent samplings in order to show that figure repetition is a common feature of these samplings.
The non-figure image in the second line second column is just a VAE artifact due to the fact that the latent distribution is {\bf not} the target 2D Gaussian, so the image is not like images in the dataset.
{\bf Second row pictures~:} results of algorithm
\ref{alg:tcalgo_sample_normal}. 			
The repetitions present in the initial i.i.d sampling (e.g. $6$, $7$, $8$, etc.) are much less present; figures never present in the first row (e.g. $3$) appear here.
%%%
{\bf Third row pictures~:} results of algorithm
\ref{alg:tcalgo_sample_empirical}.		
Results improve with respect to the second row (algorithm~\ref{alg:tcalgo_sample_normal}), only one repetition present.			
}  \label{fig:comparison}
	\end{center}
\end{figure*}

The algorithm \ref{alg:tcalgo_sample_normal} is used to sample $J=10$ points from the ideal latent distribution  $\Ncal(0_2,\textrm{Id}_2)$ (recall $L=2$); these points are then run through the decoder and we compare them with random i.i.d. sampling (plus decoder phase). We see in figure \ref{fig:comparison} that the random i.i.d. sampling has many repetitions (depending on sampling the number of repetitions may vary); on the contrary, the diversity enforcing sampling in the second row images  has fewer repetitions (a $9$ can be seen as close to a $4$ given the latent space in figure \ref{fig:2Dvisualizationmnist}; same a figure which is close to a $3$). Of course, the quality of the sampling depends on the initial VAE quality ; one component of the VAE quality is the latent distribution which, as illustrated in figure \ref{fig:2Dlatentmnist} can still be improved to match a 2D Gaussian. Since the empirical latent distribution of the dataset, depicted as the blue points in figures \ref{fig:2Dlatentmnist} and \ref{fig:2Dlatentmnist_plus_points}, does not match perfectly the target $\Ncal(0_2,\textrm{Id}_2)$ distribution, the diversity enforcing sampling,  which use $\Ncal(0_2,\textrm{Id}_2)$, will not represent an optimal sample for the empirical latent distribution ; this is seen in figure \ref{fig:2Dlatentmnist_plus_points} where the red and black points do not seem to represent optimally the blue points.

\section{Numerical results for the empirical sampling algorithm \ref{alg:tcalgo_sample_empirical}}
\label{sec:numerics_empirical}

\begin{figure*}
\begin{center}
\includegraphics[width=0.49\linewidth]{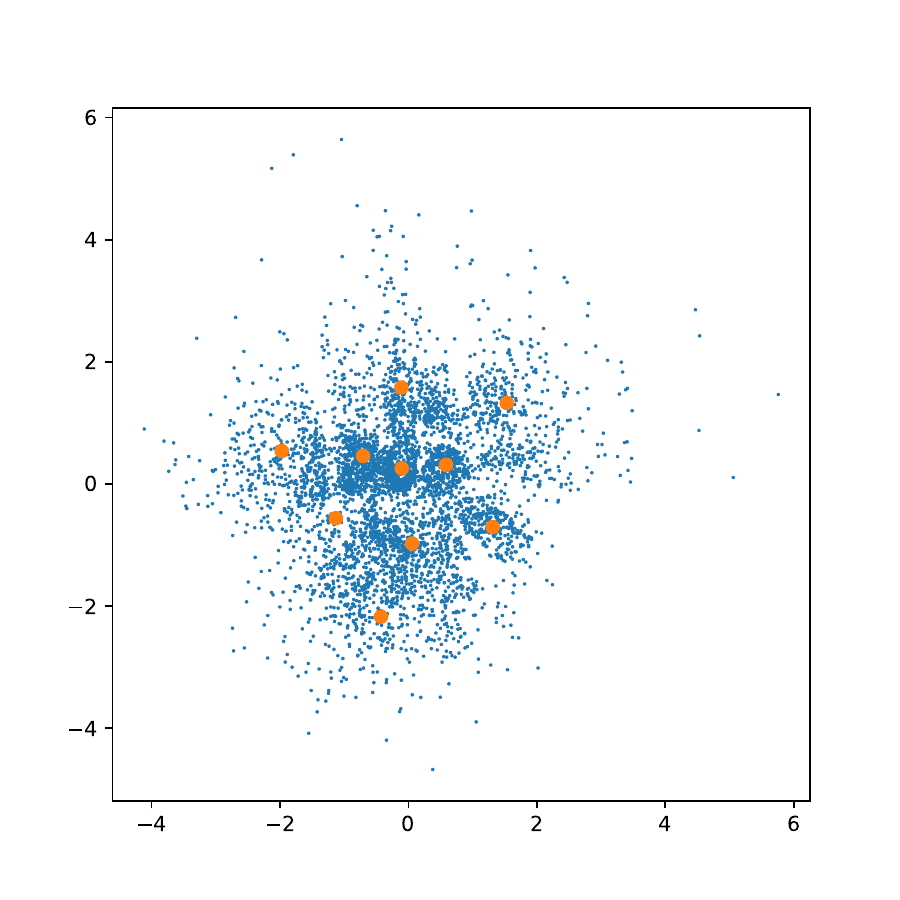}
\includegraphics[width=0.49\linewidth]{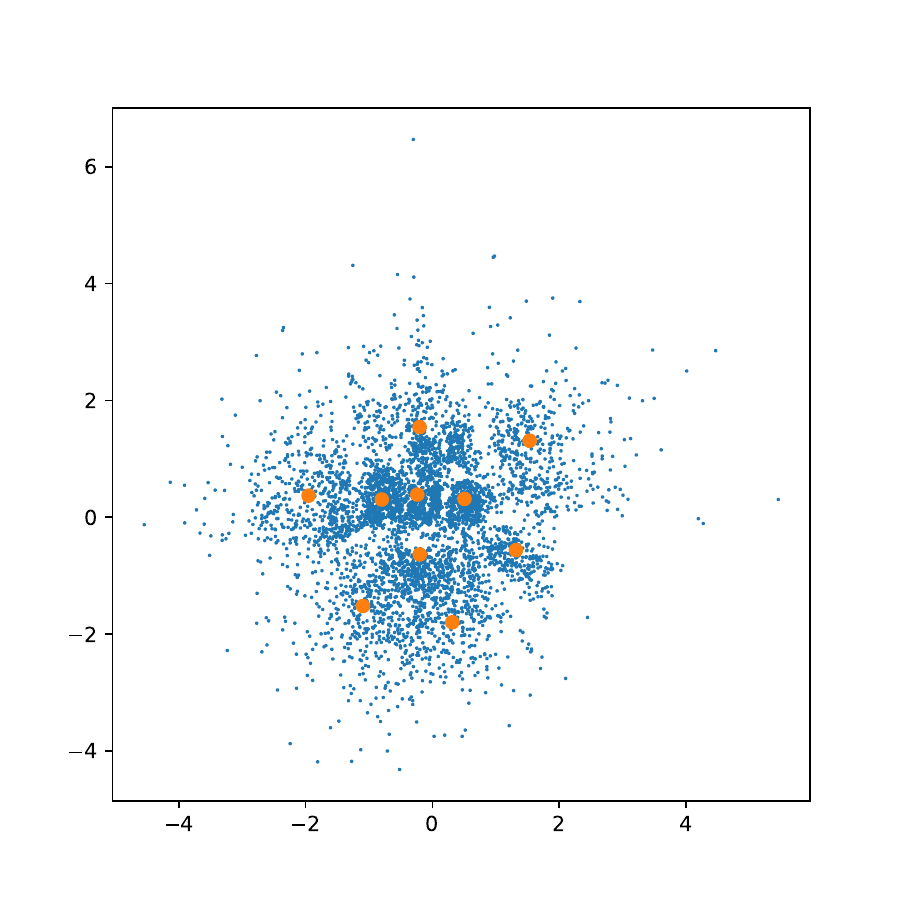}
\caption{The latent space representation of the MNIST dataset together with the two sets of latent points corresponding to diversity sampling depicted in the third row of figure
\ref{fig:comparison}. Blue points are latent distribution points $\mu_L$; orange points are the results 
	  $X=(X_j)_{j=1}^J$ of the algorithm \ref{alg:tcalgo_sample_empirical}.
}  \label{fig:2Dlatentmnist_plus_points_empirical}
\end{center}
\end{figure*}

We move now to the results for the algorithm \ref{alg:tcalgo_sample_empirical}. As indicated previously, after VAE converged we run a new epoch by asking VAE to encode all the dataset and store the latent points obtained. This was used as input for the algorithm \ref{alg:tcalgo_sample_empirical}. The VAE setting remains the same. 
The results are presented in figure \ref{fig:comparison} (third row).
The numerical results appear better than those in section \ref{sec:numerics_ideal}. This can be explained by the quality of the sampling from $\mu_L$ as illustrated in figure \ref{fig:2Dlatentmnist_plus_points_empirical} where the sampling in the latent space appear to represent more accurately the empirical distribution $\mu_L$.

\section{Discussion and final remarks}

We presented a procedure to enforce diversity in 
the decoder-based generative networks. The diversity is ensured by drawing simultaneously all samples and ensuring that the overall set is a good approximation of the target distribution on the latent space.
Two algorithms were proposed and tested numerically on standard learning datasets.
Each procedure strikes a different balance between efficiency and quality: algorithm \ref{alg:tcalgo_sample_normal} is very fast and should be used when the generative algorithm converged well and the empirical latent distribution $\mu_L$ can be supposed close to the ideal value $\Ncal(0_L,\textrm{Id}_L)$; on the other hand the algorithm \ref{alg:tcalgo_sample_empirical} 
uses the actual latent distribution $\mu_L$ and obtains better quality results but at the cost of storing $\mu_L$ (or calculating it on the fly) and can be used when the GAN / VAE / Transformer quality is not optimal. This is the one we recommend in routine practice. Nevertheless, both methods confirm the initial expectations as procedures to enhance the diversity of the  generative process.

\bibliographystyle{splncs04}
\bibliography{genbib}

\begin{thebibliography}{10}
\providecommand{\url}[1]{\texttt{#1}}
\providecommand{\urlprefix}{URL }
\providecommand{\doi}[1]{https://doi.org/#1}

\bibitem{cvae_tensorflow_jan22}
Cvae, tensorflow documentation, retrieved jan 30, 2022,
  \url{https://www.tensorflow.org/tutorials/generative/cvae}

\bibitem{allahyani_divgan_2023}
Allahyani, M., Alsulami, R., Alwafi, T., Alafif, T., Ammar, H., Sabban, S.,
  Chen, X.: {DivGAN}: {A} diversity enforcing generative adversarial network
  for mode collapse reduction. Artificial Intelligence  \textbf{317},  103863
  (2023). \doi{https://doi.org/10.1016/j.artint.2023.103863},
  \url{https://www.sciencedirect.com/science/article/pii/S0004370223000097}

\bibitem{arjovsky2017wgan}
Arjovsky, M., Chintala, S., Bottou, L.: {W}asserstein generative adversarial
  networks. In: Precup, D., Teh, Y.W. (eds.) Proceedings of the 34th
  International Conference on Machine Learning. Proceedings of Machine Learning
  Research, vol.~70, pp. 214--223. PMLR, International Convention Centre,
  Sydney, Australia (06--11 Aug 2017),
  \url{http://proceedings.mlr.press/v70/arjovsky17a.html}

\bibitem{dilokthanakul2017deep}
Dilokthanakul, N., Mediano, P.A.M., Garnelo, M., Lee, M.C.H., Salimbeni, H.,
  Arulkumaran, K., Shanahan, M.: Deep unsupervised clustering with gaussian
  mixture variational autoencoders (2017)

\bibitem{originalGAN14}
Goodfellow, I., Pouget-Abadie, J., Mirza, M., Xu, B., Warde-Farley, D., Ozair,
  S., Courville, A., Bengio, Y.: Generative adversarial nets. In: Ghahramani,
  Z., Welling, M., Cortes, C., Lawrence, N.D., Weinberger, K.Q. (eds.) Advances
  in Neural Information Processing Systems 27, pp. 2672--2680. Curran
  Associates, Inc. (2014),
  \url{http://papers.nips.cc/paper/5423-generative-adversarial-nets.pdf}

\bibitem{book_quantization_measures}
Graf, S., Luschgy, H.: Foundations of quantization for probability
  distributions. Springer (2007)

\bibitem{kingma2014adam}
Kingma, D.P., Ba, J.: Adam: A method for stochastic optimization (2014),
  arxiv:1412.6980

\bibitem{kingma_introduction_2019}
Kingma, D.P., Max, W.: An {Introduction} to {Variational} {Autoencoders}. Now
  Publishers Inc (Nov 2019)

\bibitem{kingma2013autoencoding}
Kingma, D.P., Welling, M.: {Auto-Encoding Variational Bayes} (2013),
  arxiv:1312.6114

\bibitem{Liu_2020_CVPR}
Liu, S., Wang, T., Bau, D., Zhu, J.Y., Torralba, A.: Diverse {Image}
  {Generation} via {Self}-{Conditioned} {GANs}. In: Proceedings of the
  {IEEE}/{CVF} {Conference} on {Computer} {Vision} and {Pattern} {Recognition}
  ({CVPR}). pp. 14286--14295 (Jun 2020)

\bibitem{pmlr-v119-naeem20a}
Naeem, M.F., Oh, S.J., Uh, Y., Choi, Y., Yoo, J.: Reliable fidelity and
  diversity metrics for generative models. In: III, H.D., Singh, A. (eds.)
  Proceedings of the 37th International Conference on Machine Learning.
  Proceedings of Machine Learning Research, vol.~119, pp. 7176--7185. PMLR
  (13--18 Jul 2020), \url{https://proceedings.mlr.press/v119/naeem20a.html}

\bibitem{r_gray_vector_1984}
{R. Gray}: Vector quantization. IEEE ASSP Magazine  \textbf{1}(2),  4--29 (Apr
  1984). \doi{10.1109/MASSP.1984.1162229}

\bibitem{sriperumbudur_universality_2011}
Sriperumbudur, B.K., Fukumizu, K., Lanckriet, G.R.G.: Universality,
  {Characteristic} {Kernels} and {RKHS} {Embedding} of {Measures}. Journal of
  Machine Learning Research  \textbf{12}(70),  2389--2410 (2011),
  \url{http://jmlr.org/papers/v12/sriperumbudur11a.html}

\bibitem{SZEKELY13}
Szekely, G.J., Rizzo, M.L.: Energy statistics: {A} class of statistics based on
  distances. Journal of Statistical Planning and Inference  \textbf{143}(8),
  1249--1272 (Aug 2013). \doi{10.1016/j.jspi.2013.03.018},
  \url{http://www.sciencedirect.com/science/article/pii/S0378375813000633}

\bibitem{CWAE}
Tabor, J., Knop, S., Spurek, P., Podolak, I.T., Mazur, M., Jastrz{k{e}}bski,
  S.: {Cramer-Wold AutoEncoder}. CoRR  \textbf{abs/1805.09235} (2018),
  \url{http://arxiv.org/abs/1805.09235}

\bibitem{tolstikhin2017wasserstein}
Tolstikhin, I., Bousquet, O., Gelly, S., Schoelkopf, B.: Wasserstein
  auto-encoders (2017), arxiv:1711.01558

\bibitem{turinici_radonsobolev_2021}
Turinici, G.: Radon–{Sobolev} {Variational} {Auto}-{Encoders}. Neural
  Networks  \textbf{141},  294--305 (Sep 2021).
  \doi{10.1016/j.neunet.2021.04.018},
  \url{https://www.sciencedirect.com/science/article/pii/S0893608021001556}

\bibitem{turinici_huber_2022}
Turinici, G.: Huber energy measure quantization (Dec 2022),
  \url{https://github.com/gabriel-turinici/Huber-energy-measure-quantization},
  original-date: 2022-08-25T14:07:16Z

\bibitem{turinici_supporting_2023}
TURINICI, G.: Supporting files for the paper "{Diversity} in deep generative
  models and generative {AI}", sept 2023 version (Sep 2023).
  \doi{10.5281/zenodo.7922519}, \url{https://doi.org/10.5281/zenodo.7922519}

\bibitem{vaswani_attention_2017}
Vaswani, A., Shazeer, N., Parmar, N., Uszkoreit, J., Jones, L., Gomez, A.N.,
  Kaiser, L., Polosukhin, I.: Attention {Is} {All} {You} {Need} (2017).
  \doi{10.48550/ARXIV.1706.03762}, \url{https://arxiv.org/abs/1706.03762}

\bibitem{fmnist_dataset}
Xiao, H., Rasul, K., Vollgraf, R.: Fashion-mnist: a novel image dataset for
  benchmarking machine learning algorithms. CoRR  \textbf{abs/1708.07747}
  (2017), \url{http://arxiv.org/abs/1708.07747}

\end{thebibliography}

\end{document}